%% file: air_quality_prediction_external.tex
\documentclass[sigconf, nonacm=true]{acmart}

\usepackage{multirow}
\usepackage{hyperref}

\usepackage[utf8]{inputenc} 

\AtBeginDocument{%
  \providecommand\BibTeX{{%
    \normalfont B\kern-0.5em{\scshape i\kern-0.25em b}\kern-0.8em\TeX}}}

\setcopyright{acmcopyright}
\copyrightyear{2020}
\acmYear{2020}
\acmDOI{000000000}

\acmConference[KDD '20]{ACM SIGKDD International Conference on Knowledge Discovery \& Data Mining}{August 22--27, 2020}{San Diego, CA}
\acmBooktitle{ACM SIGKDD International Conference on Knowledge Discovery \& Data Mining, August 22--27, 2020, San Diego, CA}
\acmPrice{15.00}
\acmISBN{978-1-4503-XXXX-X/18/06}

\begin{document}

\title{DeepPlume: Very High Resolution Real-Time Air Quality Mapping}







\author{Grégoire Jauvion}
\affiliation{\institution{Plume Labs}}
\email{gregoire.jauvion@plumelabs.com}

\author{Thibaut Cassard}
\affiliation{\institution{Plume Labs}}
\email{thibaut.cassard@plumelabs.com}

\author{Boris Quennehen}
\affiliation{\institution{Plume Labs}}
\email{boris.quennehen@plumelabs.com}

\author{David Lissmyr}
\affiliation{\institution{Plume Labs}}
\email{david.lissmyr@plumelabs.com}


\begin{abstract}
This paper presents an engine able to predict jointly the real-time concentration of the main pollutants harming people's health: nitrogen dioxyde (NO$_{2}$), ozone (O$_{3}$) and particulate matter (PM$_{2.5}$ and PM$_{10}$, which are respectively the particles whose size are below $2.5~\mu m$ and $10~\mu m$).

The engine covers a large part of the world and is fed with real-time official stations measures, atmospheric models' forecasts, land cover data, road networks and traffic estimates to produce predictions with a very high resolution in the range of a few dozens of meters. This resolution makes the engine adapted to very innovative applications like street-level air quality mapping or air quality adjusted routing.

Plume Labs has deployed a similar prediction engine to build several products aiming at providing air quality data to individuals and businesses. For the sake of clarity and reproducibility, the engine presented here has been built specifically for this paper and differs quite significantly from the one used in Plume Labs' products. A major difference is in the data sources feeding the engine: in particular, this prediction engine does not include mobile sensors measurements.
\end{abstract}




\maketitle

\input{content_external}


\bibliographystyle{ACM-Reference-Format}
\bibliography{air_quality_prediction_external}










\end{document}

%% file: content_external.tex
\section{Introduction}

Air pollution is one of the major public health concern. World Health Organization (WHO) estimates that more than 80\% of citizens living in urban environments where air quality is monitored are exposed to air quality levels that exceed WHO guideline limits. It also estimates that $4.2$ million deaths every year are linked to outdoor air pollution \cite{WHO_report}.

Despite those alarming figures, very few citizens have access to information about the quality of the air they breathe. More and more public and private initiatives are being developed to close this gap and give to citizens the information they need to protect themselves from air pollution.

This is a particularly challenging topic because air quality varies a lot, both in time and in space. For example, a polluted air can clean itself in a few hours after a heavy rain. Also, a crowded street can be much more polluted than a green park area a few hundreds meters away \cite{AQ_review}.

One of the key difficulties when it comes to air quality modeling is the lack of data: it is believed that there are about $30$ thousands air quality monitoring stations worldwide, which is orders of magnitude below the number of stations needed given how much air quality varies in space. Also, there is as far as we know no comprehensive database providing the monitoring stations live and historical measurements, and building this database is a very time-consuming task.

The quantity of air quality data varies a lot depending on the location: urban areas in developed countries are generally well monitored, with dozens of monitoring stations in cities like London or Beijing. However, rural areas and poorer countries may be covered very sparsely given the high cost of building and maintaining an air quality monitoring network (the monitoring stations set up by public authorities cost generally between \$10k and \$50k per station).

This paper presents an engine able to predict the concentrations of atmospheric pollutants regulated by WHO: nitrogen dioxide (NO$_{2}$), ozone (O$_{3}$) and particulate matter (PM$_{2.5}$ and PM$_{10}$, which are respectively the particles whose size are below $2.5~\mu m$ and $10~\mu m$). The predictions are performed at a large scale covering whole countries or continents and with a high resolution in the range of a few dozens of meters.

\begin{figure}[h]
  \centering
  \includegraphics[width=\linewidth]{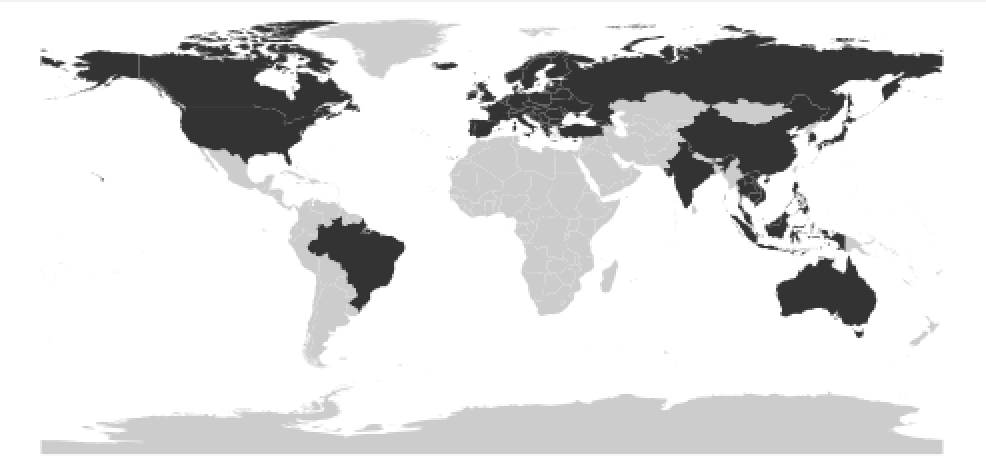}
  \caption{Map of the regions covered by the prediction engine (in black)}
  \label{fig:geo_cover}
  \Description{The regions covered by the prediction engine are displayed in black on this map.}
\end{figure}

Air quality depends highly on the region considered: for example, coal power plants produce a large part of air pollution in countries like China and India \cite{power_plant_emissions} while their impact is more limited in Western Europe. For this reason, we choose to estimate prediction models region per region. The geographical cover of the prediction engine is shown on Figure~\ref{fig:geo_cover}.

The engine uses the air quality measurements provided by thousands of monitoring stations across the world, atmospheric simulations run by atmospheric science labs whose results are made available, and anthropogenic emissions estimates based on diverse datasets.

It is based on a somewhat classical neural network architecture, and tackles the data sparsity issue encountered in some regions with a transfer learning strategy: a global prediction model is learnt on a global dataset, and the final prediction layer is specialized on each regional dataset.

The paper is organized as follows. We discuss earlier works in Section 2. Section 3 gives a detailed overview of the data sources used, and describes how the features feeding the engine are built. Section 4 presents the structure of the prediction engine and details the model estimation process. Section 5 provides an evaluation of the prediction engine. Section 6 lists a few applications built on top of it.

\section{Related work}

The problem of air quality prediction is much studied in the literature and is tackled through various angles. \cite{deep_learning_review} and \cite{bigdata_summary} present comprehensive reviews of air quality modeling using machine learning approaches.

Some papers focus on air quality temporal forecasts and aim at predicting pollutants concentration at air quality monitoring stations using the stations historical measurements and meteorological features. Most of them are based on neural networks (generally RNN or LSTM architectures) and the forecast horizon varies from a few hours to $48$ hours. \cite{deepair}, \cite{china_deep_rnn} and \cite{deep_cnn_lstm} build forecast models in China, \cite{deep_air_net} and \cite{india_lstm_rnn} build models for indian cities.

Other papers propose spatiotemporal modeling frameworks. \cite{conv_lstm} and \cite{conv_lstm_2} use Convolutional LSTM networks introduced in \cite{conv_lstm_theory} for precipitation forecasting. The networks are fed with monitoring stations measurements and meteorological features to build 24 hours to 48 hours air quality forecasts. The spatial resolution in \cite{conv_lstm} is $0.1$ degree. \cite{airnet}, \cite{deep_air_learning} and \cite{deep_air_quality} use similar deep learning architectures to build air quality forecasts in China. \cite{airnet} covers all China with a $0.25$ degree resolution, \cite{deep_air_learning} covers Beijing with a $1$ kilometer resolution and \cite{spatial_temporal} focuses also on Beijing. \cite{deep_distributed} provides fine-grained forecasts in $300$ Chinese cities using a deep learning spatio-temporal architecture.

\cite{u-air} and \cite{fine-grained} propose different architectures mixing a spatial and a temporal predictor to build air quality forecasts in a few chinese cities with a $1$ kilometer resolution. Their approach uses additional features like traffic or land-use features.

Also, some papers introduce equations based on dispersion modeling in the prediction methodology in order to reduce overfitting due to the low number of monitoring stations available. \cite{physical_model} uses a hybrid architecture to build air quality maps in Tianjin, China, and \cite{model_data_driven} builds air quality forecasts in Chengdu, China.

\cite{global_approach} and \cite{r_package_airpred} focus on spatial air quality predictions. \cite{global_approach} builds global PM2.5 predictions based on monitoring stations and satellite-based measurements with a $0.1$ degree resolution using statistical modeling. \cite{r_package_airpred} models PM2.5 in the US using monitoring stations measurements, atmospheric models outputs and land-use datasets.

Finally, a few papers show how using mobile low-cost sensors networks can improve air quality predictions accuracy. \cite{aclima_mapping} shows the results of experiments in Mountain View and San Francisco, \cite{mobile_network} presents modeling approaches on an experimental setup in Lausanne, Switzerland, and \cite{china_maps} presents a case-study in Beijing using mobile sensors. These approaches are very promising but need expensive setups (deployment of mobile sensors networks) which can not be performed on a global scale.

\section{Data sources and features}

This section details the data sources used by the prediction engine and the features computation for each data source. The availability and reliability of those data sources vary from a region to the other and are a key factor of the engine's overall accuracy.

We introduce the euclidean distance $\| l - l' \|$ between two locations $l$ and $l'$. We define also the exponential kernel $k_d(l, l') = exp(-\frac{\| l - l' \|}{d})$, where the distance $d$ is expressed in kilometers.

\subsection{Air quality data}

\subsubsection{Monitoring stations measurements}

\begin{figure}[h]
  \centering
  \includegraphics[width=\linewidth]{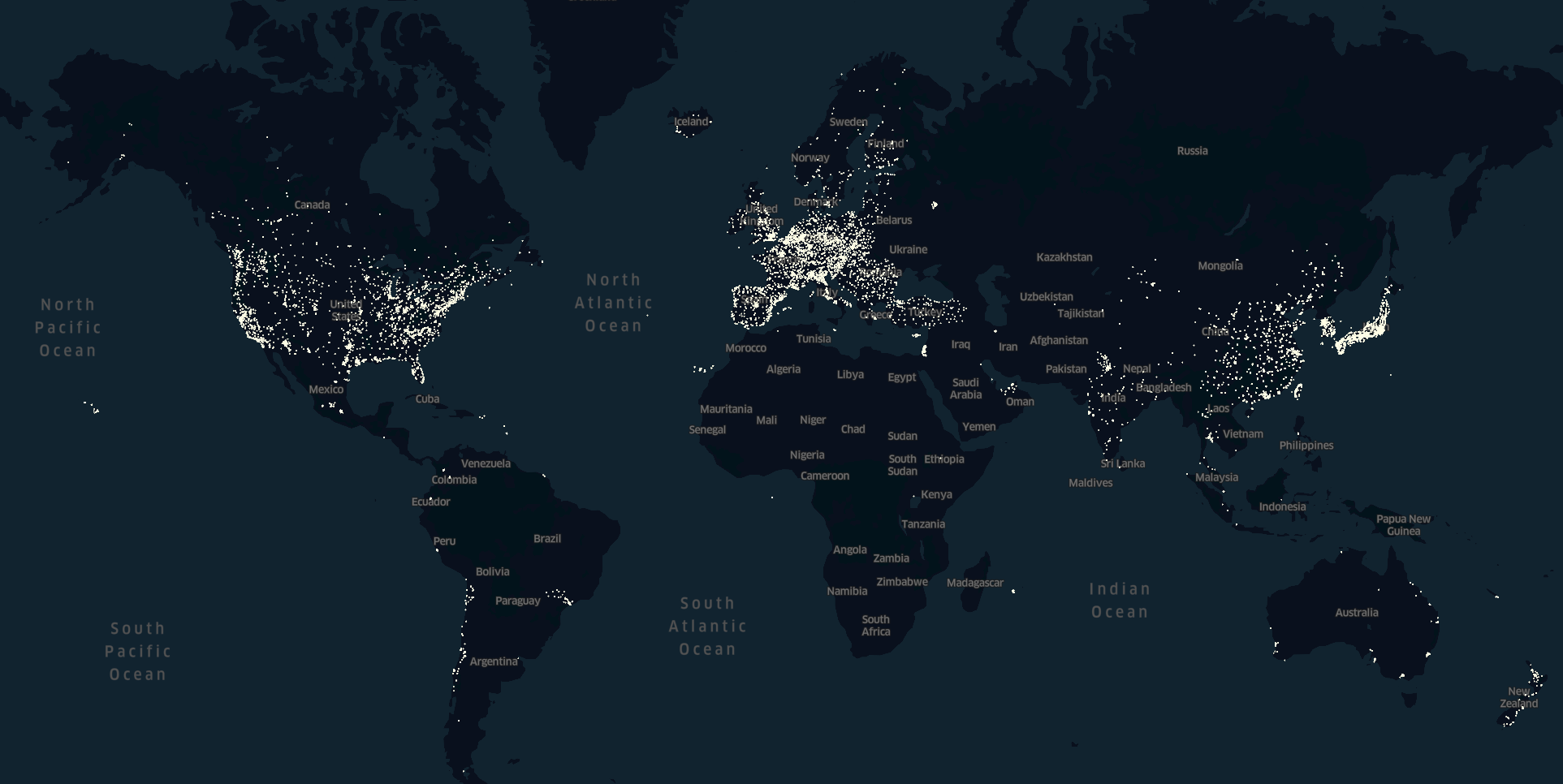}
  \caption{Map of the monitoring stations whose measurements are included in the predictions}
  \label{fig:stations}
  \Description{This map shows the locations of the monitoring stations whose air quality measurements feed the prediction engine.}
\end{figure}

We have built a proprietary architecture based on several dozens of crawlers collecting the air quality measurements provided by about $14000$ monitoring stations across the world. Figure~\ref{fig:stations} shows a global map of the locations of those monitoring stations.

Each monitoring station does not necessarily measure the four pollutants predicted by the engine. The resulting missing measurements have been flagged as $NA$ in the datasets and are treated specifically. Table~\ref{tab:stations} gives the number of monitoring stations per region as well as the number of stations measuring each pollutant. As almost all monitoring stations give measurements on a hourly basis, the datasets have been built on a hourly basis.

\begin{table}
  \caption{Number of monitoring stations per region and pollutant}
  \label{tab:stations}
  \begin{tabular}{c||c|cccc}
    \toprule
    Region & Global & NO2 & O3 & PM2.5 & PM10 \\
    \midrule
    Europe & 2778 & 2252 & 1614 & 790 & 1958 \\
    United States & 1924 & 331 & 1300 & 1041 & 347 \\
    Japan / S. Korea & 1924 & 1804 & 309 & 1345 & 313 \\
    China & 1390 & 1385 & 1387 & 1387 & 1387 \\
    Canada & 299 & 103 & 236 & 220 & 17 \\
    Southeast Asia & 66 & 55 & 52 & 41 & 44 \\
    India & 231 & 223 & 213 & 224 & 201 \\
    Brazil & 51 & 39 & 41 & 20 & 46 \\
    Australia & 66 & 42 & 34 & 47 & 54 \\
    Russia & 29 & 20 & 15 & 15 & 21 \\
  \bottomrule
\end{tabular}
\end{table}

We include in the datasets the measurements from January 1st, 2018 to December 31st, 2019. We have found that there are missing and erroneous values (generally abnormally high) coming from those monitoring stations. They can be encountered during station maintenance windows, during station failures or if issues arise during the publishing or collection of said data. While we are not able to determine the exact cause of such errors, it is important to detect them and define an appropriate treatment: missing values are discarded from the datasets, and erroneous values are detected using an outlier detection engine and then discarded.

For any distance $d$ and for each pollutant $p$, we define the feature $\text{StationsMeasures}_{t,d,p}$ at every location $l$ as the weighted average of the monitoring stations measurements for pollutant $p$ at time $t$. We define also $\text{StationsCounters}_{t,d,p}$ as the sum of the weights. The weights are computed with an exponential kernel $k_d$.

$\text{StationsMeasures}_{t,d,p}$ can be interpreted as a predictor of the concentration of pollutant $p$ based on the measurements of the monitoring stations around. $\text{StationsCounters}_{t,d,p}$ measures the density of monitoring stations and can be interpreted as a level of confidence in $\text{StationsMeasures}_{t,d,p}$: the more monitoring stations there are around a point, the more robust the weighted average measure is.

\subsubsection{Mobile sensors data}

While stations measurements provide useful ground-truth values for air quality modeling, such data sources are constrained by two limits:
\begin{itemize}
  \item Data sparsity: because monitoring stations are expensive to build and maintain, large areas are not or sparsely covered, including in densely populated areas
  \item Data delay: there is often a significant delay between the time a measurement is made and the time this measurement is made available. This delay can be as high as a few hours, thus decreasing the relevance of the measurement for live monitoring and forecasts
\end{itemize}

Mobile low-cost sensors provide an abundant source of data, and unlike fixed stations which monitor a single location, they are able to map a larger area. We therefore believe that those sensors are a valuable and useful complement to standard monitoring stations. To that end, Plume Labs has developed Flow\footnote{\url{https://plumelabs.com/en/flow/}}, a personal wearable air quality sensor that measures NO$_{2}$, VOCs, PM$_{10}$, PM$_{2.5}$ and PM$_{1}$ on a minute-by-minute basis. This sensor being cheaper and easier to maintain than typical monitoring stations, it can be deployed in ad-hoc sensing networks over selected regions to provide prediction engines with valuable ground-truth data.

The use of both fixed stations and low-cost mobile sensors measurements in the prediction engine is a particularly challenging area of research. Gaussian process regression provides a natual theoretical framework to aggregate both sources of data in a consistent way, taking into account the different accuracy of the measurements coming from fixed stations and low-cost sensors.

\subsubsection{Atmospheric models simulations}

Atmospheric models rely on physical and chemical modeling of pollutants' emissions and dispersion, and the simulations are initiated with monitoring stations and satellite-based measurements. Each atmospheric model is characterized by its geographical cover (regional or global) and its spatial granularity (generally a few dozens of kilometers).

We define the feature $\text{AtmosphericModel}_{t,p}$ at every location $l$ by doing a bilinear interpolation of the forecasts at time $t$ produced on the grid for pollutant $p$ by the atmospheric model with the best resolution around $l$.

It is worth noting that those simulations have a poor spatial granularity compared to the resolution reached by our engine. However, they provide a very valuable input in regions which are not or sparsely covered with monitoring stations, typically outside urban areas.

\subsection{Anthropogenic emissions datasets}

\subsubsection{Traffic data}

We collect traffic data through a real-time \textit{jam factor} over each road segment of a given area. The \textit{jam factor} is a value between $0$ and $10$ measuring the road congestion: $0$ means that the road is not congested while $10$ means that it is very congested. Historical and real-time traffic data are collected across Europe and the United States.

For any distance $d$, we define the feature $\text{Traffic}_{t,d}$ at every location $l$ as the weighted sum of the jam factors of the road segments around location $l$, and each road segment is weighted by the product of its length and its functional class\footnote{The functional class is a classification of each road segment, from 1 (meaning a small road) to 5 (meaning a large road).}. The weights are computed with the exponential kernel $k_d$.

The product of the jam factor, the length and the functional class on a road segment is supposed to be proportional to the traffic emissions on this road segment. Hence, this feature can be interpreted as a proxy of traffic emissions around.

\subsubsection{Road networks}

We have road network details and topology in the regions covered by the prediction engine. Road network data is collected as a set of road segments, and any road segment is associated with a set of metadata regrouping a significant number of informations including a classification per usage, e.g., \textit{motorway} or \textit{residential}. We have built two aggregate categories named \text{Roads} and \text{MajorRoads}.

We define the features $\text{Roads}_d$ and $\text{MajorRoads}_d$ for every location $l$ as the weighted length of roads and major roads around location $l$, the weights being computed with an exponential kernel $k_d$. Those features estimate the roads density around a location $l$, and can be interpreted as a rough proxy of the number of vehicles around.

\subsubsection{Land-use data}

Land-use datasets (also called land cover datasets), classify the land into several categories like residential areas or green areas. Different datasets have been used, each characterized by its geographical cover, its spatial resolution and the granularity of the categories provided. We have built three aggregate categories named \text{Industry}, \text{Residential} and \text{Green}.

For each category, we define the corresponding feature for a distance $d$ at every location $l$ as the weighted share of land flagged with the category. The weights are computed with the exponential kernel $k_d$.

\subsubsection{Power plants emissions}

Electrical power plants, and in particular coal power plants are a major source of air pollution in some regions. A global power plant database giving the location, maximum capacity and type (coal, gas or oil) of all power plants is used to build an estimation of the resulting pollution emissions.

\section{Model estimation}

The prediction engine maps the features with the 4-dimensional vector giving the pollutants concentrations in $\mu{}g/m^{3}$ using a classical neural network with $2$ hidden layers of sizes $n_1$ and $n_2$. We use rectified linear activation functions in the inner layers of the network, and the identity function in the output layer.

\subsection{Datasets' description}

In each region, we have built a dataset containing the monitoring stations measurements as well as the features used by the prediction engine. Each data point corresponds to the measurements returned by an air quality monitoring station at a given time. The time range covers a period from January 1st, 2018 to December 31st, 2019. We have considered for each feature different distances $d$ from $50$ meters to $100$ kilometers. Missing measurements are flagged as $NA$: most of the time, it is because the monitoring station does not monitor the pollutant considered. The datasets have been built in Python using various packages including Scikit-Learn and Numpy. Table~\ref{tab:dataset} gives the total number of data points in each region.

It is worth noting that for each data point, the corresponding monitoring station is excluded from the monitoring stations used to build the features $\text{StationsMeasures}_{t,d,p}$ and $\text{StationsCounters}_{t,d,p}$, to make sure that the measurements on which the predictor is trained are not included in the features.

\begin{table}
  \caption{Number of data points per region}
  \label{tab:dataset}
  \begin{tabular}{c|c}
    \toprule
    Region & Number of data points \\
    \midrule
    Europe & 36.2 mlns \\
    United States & 26.9 mlns \\
    Japan / S. Korea & 29.0 mlns \\
    China & 18.9 mlns \\
    Canada & 4.29 mlns \\
    Southeast Asia & 0.88 mlns \\
    India & 0.82 mlns \\
    Brazil & 0.67 mlns \\
    Australia & 0.61 mlns \\
    Russia & 0.27 mlns \\
  \bottomrule
\end{tabular}
\end{table}

\subsection{Estimation setup}

In each region, the set of monitoring stations is splitted randomly into two parts: $80\%$ of the stations are used to build the training dataset and the remaining $20\%$ are used to build the evaluation dataset. This method ensures that the evaluation is performed at locations which are not included at all during the training.

Then, the training and evaluation datasets are built by sampling the data points the following way:
\begin{itemize}
  \item Plume Labs internal Air Quality Index\footnote{An Air Quality Index~(AQIs) is a normalization of a pollutant raw concentration in $\mu{}g/m^{3}$ to a health impact scale, allowing inter-pollutant comparison. AQIs are commonly used worldwide and are usually defined locally at the country or continent scale to comply with local standards. Plume Labs AQI is based on WHO recommendations and is extensively described here: \url{https://plumelabs.zendesk.com/hc/en-us/articles/360008268434-What-is-the-Plume-AQI-}. A global AQI value can be calculated as the maximum of single pollutants' AQIs.}~(PAQI) (see \cite{plume_aqi}) is computed for each data point (using the pollutants with non-missing measurements only), and is used to split the datasets into $4$ parts depending on the corresponding air pollution exposure: Low, Moderate, High and Very High
  \item For each region, the same number of data points are sampled with replacement in each exposure category. In the experiments detailed in this paper, we have sampled $2$ million points in each exposure category to build the training dataset, and $200$ thousand points in each exposure category to build the evaluation dataset. Consequently, for each region, the size of the training dataset is $8$ million points, and the size of the evaluation dataset is $800$ thousand points
\end{itemize}

This sampling ensures that the prediction engine remains accurate for all air quality levels, while using the raw datasets leads to a better accuracy for good air quality levels at the cost of a worse accuracy when pollution levels are high, because low pollutants concentrations are more frequent in the datasets.

Then, the models are trained on the training datasets using TensorFlow and Keras. The loss used in training is the mean squared logarithmic error (MSLE) loss. $NA$ values are excluded of the loss computation. We use Adam optimizer with a learning rate equal to $0.001$. The batch size is $1024$. In Europe and in the United States, the prediction models are estimated on the Europe and United States datasets respectively. In the other regions, we use a transfer learning strategy described below.

\subsection{Transfer learning strategy in regions with few data}

\begin{figure}[h]
  \centering
  \includegraphics[width=\linewidth]{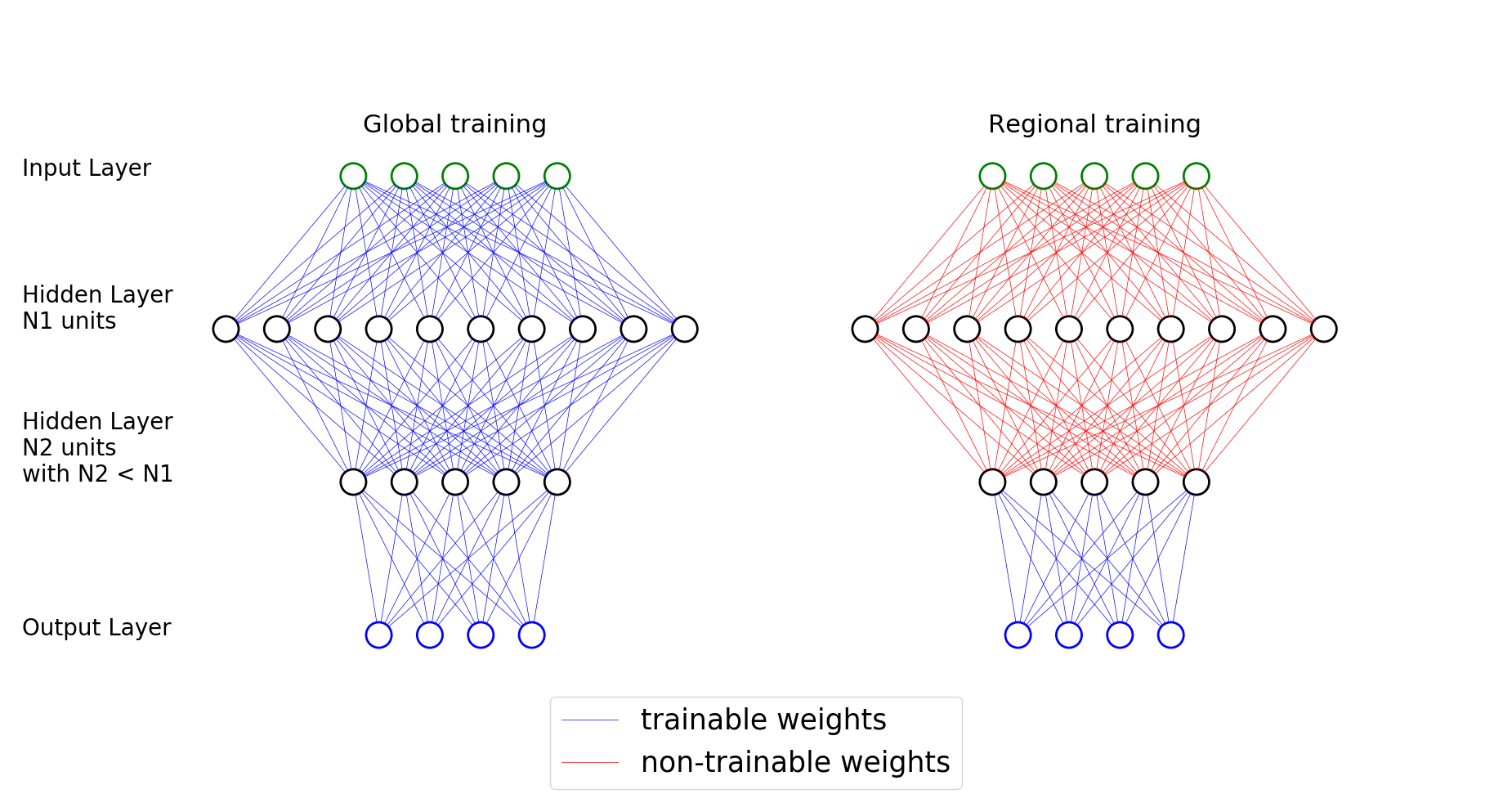}
  \caption{Transfer learning strategy}
  \label{fig:transfer_learning_diagram}
  \Description{This diagram highlights the transfer learning strategy applied in regions with few data.}
\end{figure}

In some regions where there are few monitoring stations, the prediction model trained using the monitoring stations located in the region only was not robust enough. We use the following transfer learning strategy, summarized on Figure~\ref{fig:transfer_learning_diagram}:
\begin{itemize}
  \item First, a global model is learnt on a global dataset built by concatenating the datasets of all the regions. This phase corresponds to the left network on Figure~\ref{fig:transfer_learning_diagram}
  \item Then, the final prediction layer of the global model is trained on the regional dataset. This phase corresponds to the right network on Figure~\ref{fig:transfer_learning_diagram}
\end{itemize}

This strategy has been applied in the regions where the datasets contain less than $1$ million data points: this threshold has been determined empirically. The next section shows how it improves the predictions accuracy in those regions.

\subsection{Features selection}

The results presented in this paper are based on a prediction model using a subset of the features described in Section 3. In particular, mobile sensors measurements are not integrated. Table~\ref{tab:features} summarizes the features used by the prediction engine in each region.

\begin{table}
  \caption{Features used in each region}
  \label{tab:features}
  \begin{tabular}{cc}
    \toprule
    Region & Features \\
    \toprule
    \multirow{7}{2cm}{Europe, United States} & $\text{StationsMeasures}_{t,1,p}$ for all pollutants,\\
                          & $\text{StationsMeasures}_{t,10,p}$ for all pollutants,\\
                          & $\text{StationsMeasures}_{t,100,p}$ for all pollutants,\\
                          & $\text{StationsCounters}_{t,10,p}$ for all pollutants,\\
                          & $\text{AtmosphericModel}_{t,p}$ for all pollutants,\\
                          & $\text{Traffic}_{t,0.1}$, $\text{Roads}_{0.1}$, $\text{MajorRoads}_{0.1}$,\\
                          & $\text{Industry}_{0.1}$, $\text{Residential}_{0.1}$, $\text{Green}_{0.1}$\\
    \midrule
    \multirow{7}{2cm}{Other regions} & $\text{StationsMeasures}_{t,1,p}$ for all pollutants,\\
                                    & $\text{StationsMeasures}_{t,10,p}$ for all pollutants,\\
                                    & $\text{StationsMeasures}_{t,100,p}$ for all pollutants,\\
                                    & $\text{StationsCounters}_{t,10,p}$,\\
                                    & $\text{AtmosphericModel}_{t,p}$ for all pollutants\\
                                    & $\text{Roads}_{0.1}$, $\text{MajorRoads}_{0.1}$,\\
                                    & $\text{Residential}_{0.1}$, $\text{Green}_{0.1}$ \\
  \bottomrule
\end{tabular}
\end{table}

We can give the following interpretations for the features selected:
\begin{itemize}
  \item The features $\text{StationsMeasures}_{t,d,p}$ can be interpreted as an estimate of air quality based on the monitoring stations around. The distances $1$, $10$ and $100$ kilometers have been selected: $d=1$ is a good predictor  in areas with many monitoring stations while higher distances ($d=10$ and $d=100$) are more relevant when the monitoring stations are sparsely located. Using distances higher than $100$ kilometers did not improve the accuracy of the predictions
  \item The counters $\text{StationsCounters}_{t,10,p}$ provide a proxy of the surrounding stations' density, which can be interpreted as the robustness of the weighted average of the stations measurements. Adding the counters computed with other distances did not bring additional predictive power at the cost of adding several features: that is why one distance only is used
  \item The feature $\text{AtmosphericModel}_{t,p}$ based on the atmospheric models forecasts can be interpreted as a background air quality level and adds predictive power in regions with few monitoring stations
  \item The features modeling the impact of human activity (land cover, roads network and traffic) are built with a much smaller distance ($100$ meters) which brings much more predictive power than higher distances
\end{itemize}

\section{Model evaluation}

\subsection{Prediction model evaluation}

The prediction models estimated for each region are compared to a benchmark predictor which consists in predicting for each pollutant the measure provided by the closest monitoring station. The evaluation metric is the MSLE loss function used to train the models. Table~\ref{tab:evaluation} compares the metric computed for the prediction model and the benchmark predictor on the evaluation dataset for each region. The model brings a very significant MSLE decrease in all regions compared to the benchmark predictor.

\begin{table*}
  \caption{MSLE computed on the evaluation dataset}
  \label{tab:evaluation}
  \begin{tabular}{c||cc|c}
    \toprule
    Region & Benchmark & Prediction model & Improvement (in \%) \\
    \midrule
    Europe  & 0.355 & 0.184 & -48.2 \\
    United States  & 0.461 & 0.318 & -31.0 \\
    Japan / S. Korea  & 0.233 & 0.14 & -39.9 \\ 
    China  & 0.317 & 0.214 & -32.5 \\
    Canada  & 0.441 & 0.392 & -11.1 \\
    Southeast Asia  & 1.234 & 0.901 & -27.0 \\
    India  & 0.698 & 0.351 & -49.7 \\
    Brazil  & 0.442 & 0.318 & -28.1 \\
    Australia  & 0.355 & 0.305 & -14.1 \\
    Russia  & 0.541 & 0.387 & -28.5 \\
  \bottomrule
\end{tabular}
\end{table*}

\begin{figure*}[h]
  \centering

  \includegraphics[width=.45\linewidth]{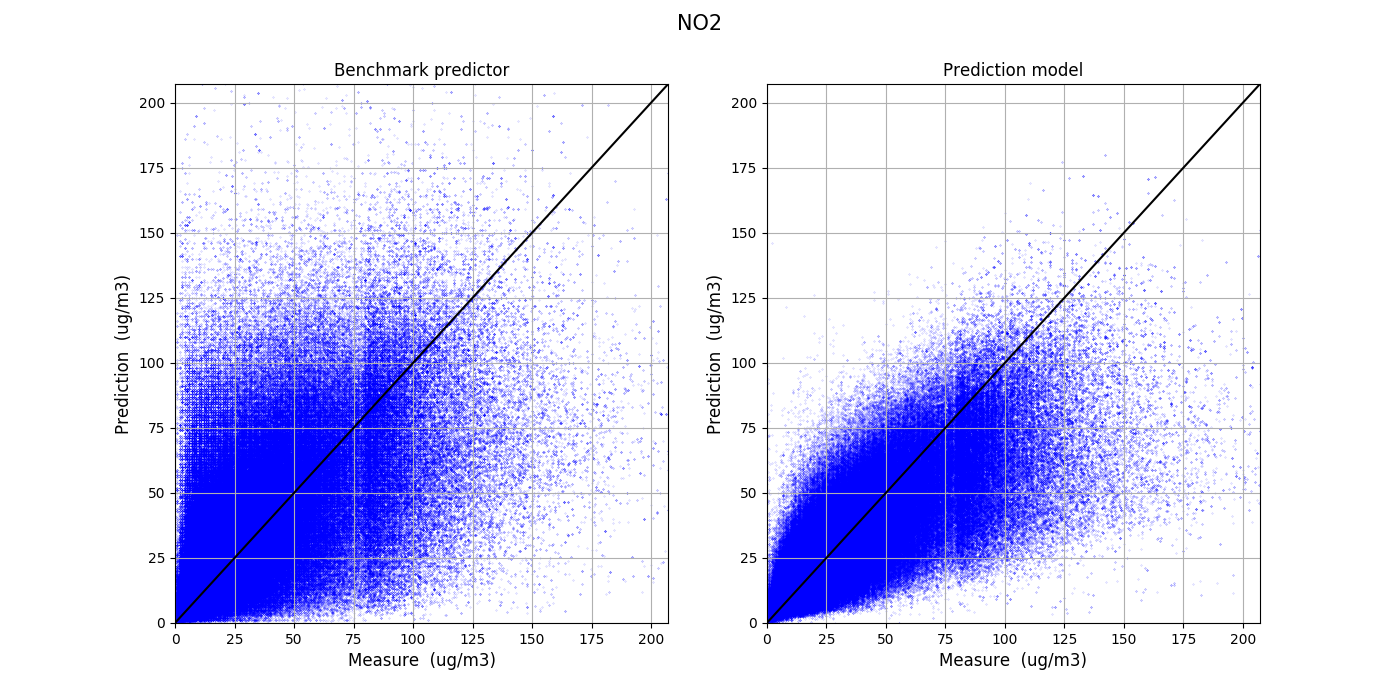}
  \includegraphics[width=.45\linewidth]{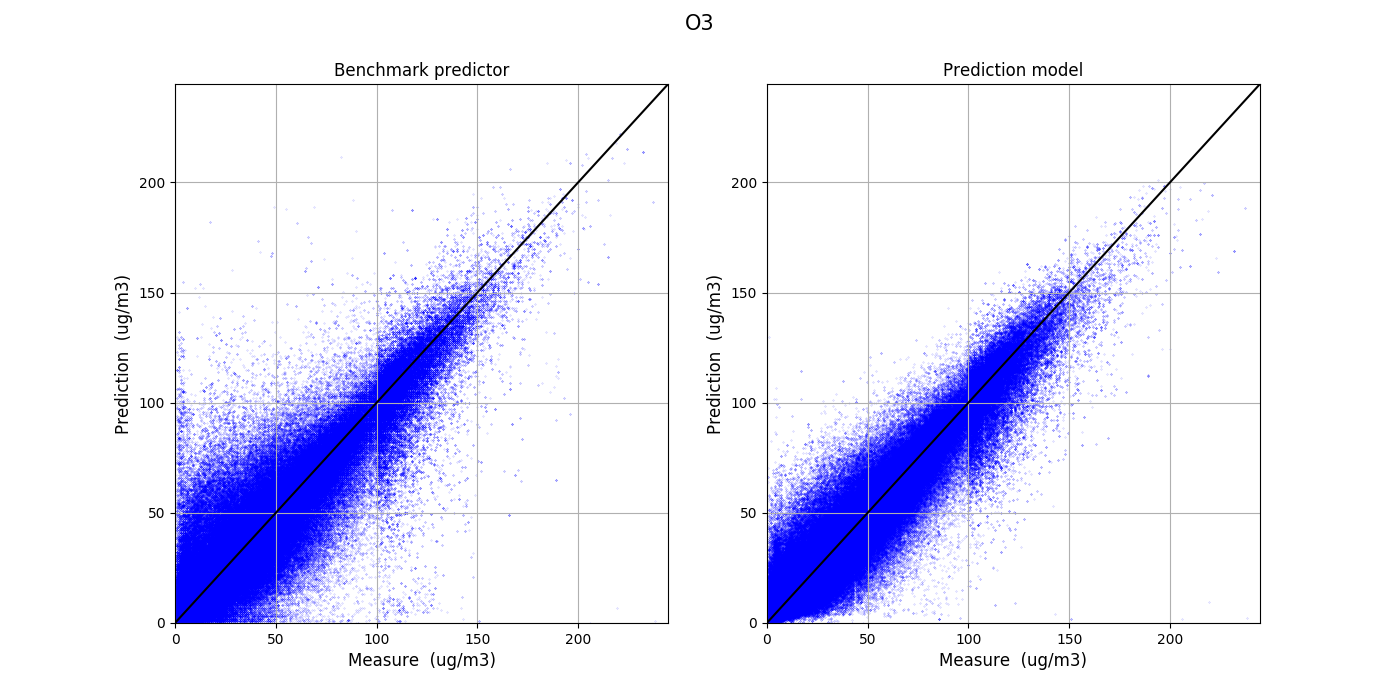}
  \includegraphics[width=.45\linewidth]{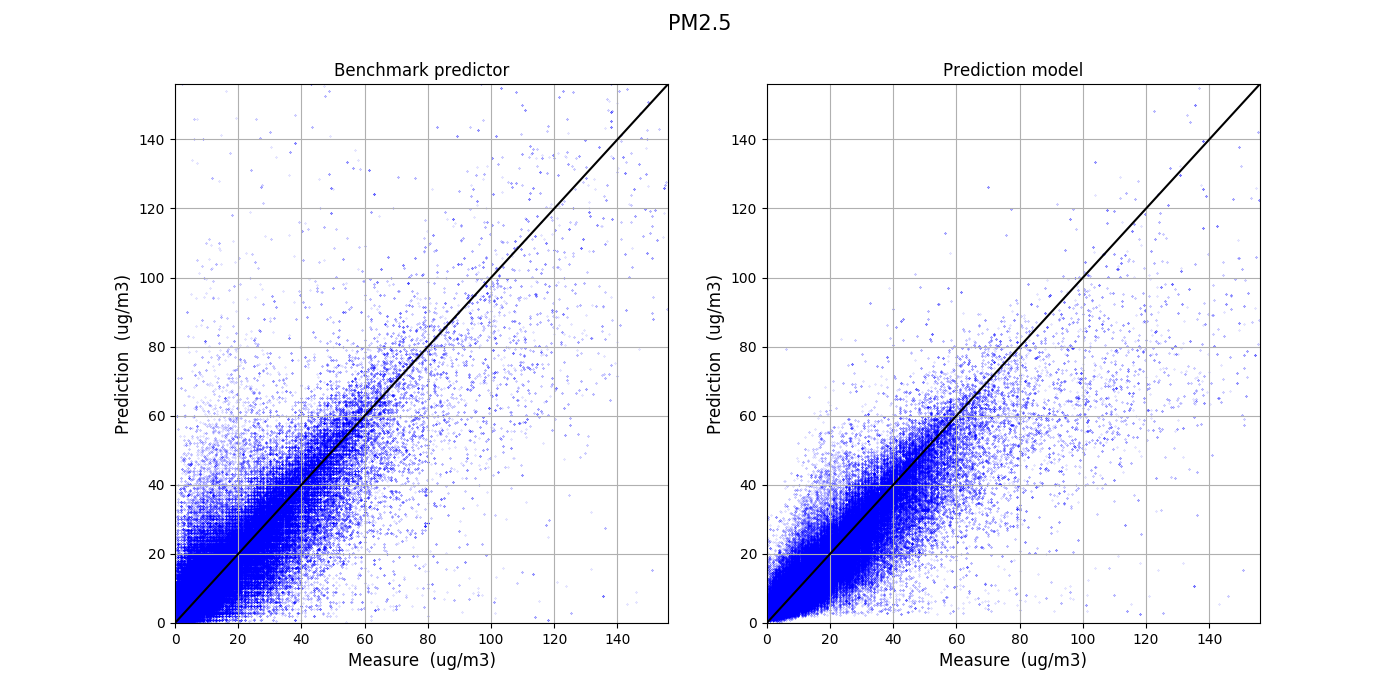}
  \includegraphics[width=.45\linewidth]{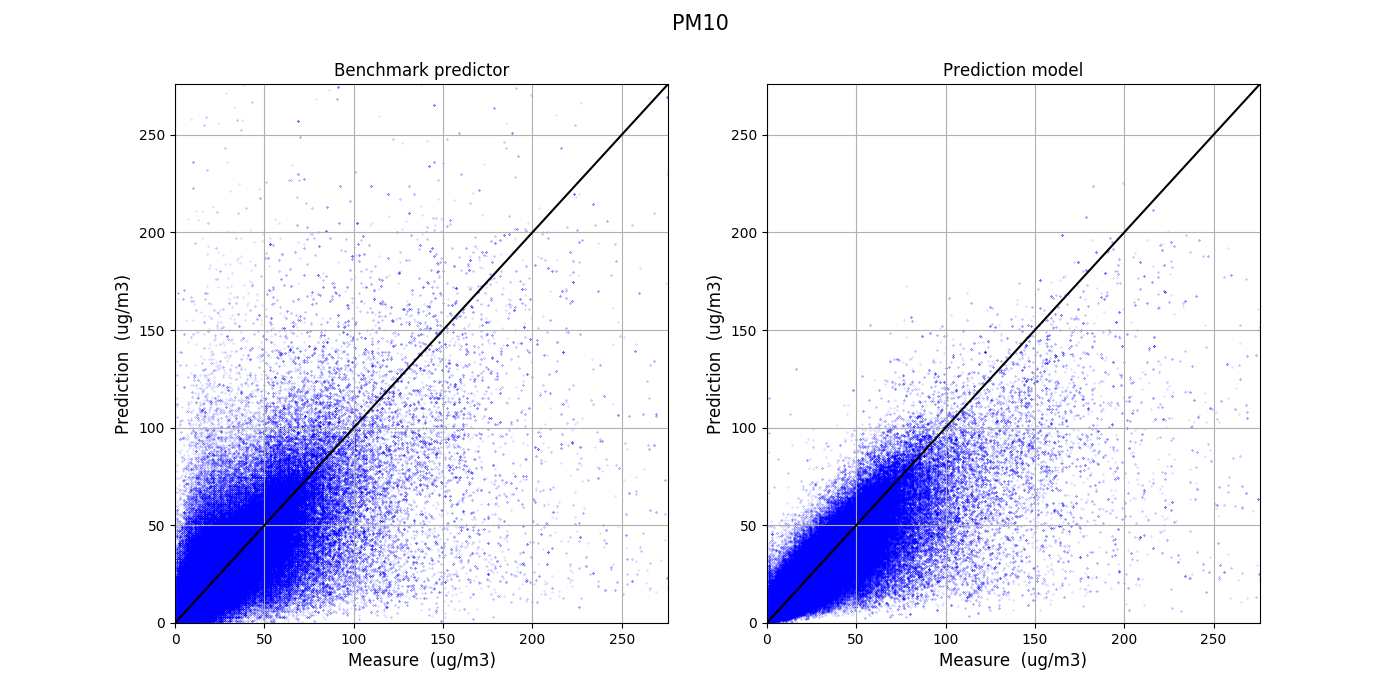}

  \caption{For each pollutant, these plots show the air quality measurements (X axis) and the predictions obtained by the benchmark predictor (Y axis, left plot) or the prediction model (Y axis, right plot)}
  \label{fig:prediction}
  \Description{For each pollutant, these plots show the air quality measurements (X axis) and the predictions obtained by the benchmark predictor (Y axis, left plot) or the prediction model (Y axis, right plot)}
\end{figure*}

Europe is the region with the most monitoring stations and consequently the biggest dataset. On Europe evaluation dataset we present a more detailed evaluation:
\begin{itemize}
  \item Table~\ref{tab:evaluation_europe} gives the MSLE computed for each pollutant for the prediction model and the benchmark predictor. It shows that the prediction model performs significantly better than the benchmark predictor for every pollutant
  \item Figure~\ref{fig:prediction} displays for each pollutant the measure and the concentration predicted by the prediction model or the benchmark predictor on the evaluation dataset. The predictions obtained with the prediction model are clearly much closer to the actual measurements than the ones produced by the benchmark predictor
\end{itemize}

\begin{table}
  \caption{Europe: MSLE computed per pollutant on the evaluation dataset}
  \label{tab:evaluation_europe}
  \begin{tabular}{c||cc|c}
    \toprule
    Pollutant & Benchmark & Prediction model & Improvement (in \%) \\
    \midrule
    NO2 & 0.530 & 0.230 & -56.6 \\
    O3 & 0.330 & 0.215 & -34.8 \\
    PM2.5 & 0.289 & 0.154 & -46.7 \\
    PM10 & 0.271 & 0.139 & -48.7 \\
  \bottomrule
\end{tabular}
\end{table}

\subsection{Evaluation of the transfer learning strategy}

In the regions where transfer learning is used to estimate the prediction models, we compare the accuracy of the three following models: the model trained on the global training dataset (Global), the model trained on the regional training dataset only (Regional), and the model trained using the transfer learning approach (Transfer learning). Table ~\ref{tab:transfer_evaluation} compares the evaluation metric computed on each region's evaluation dataset with those $3$ models. In all regions excluding Brazil, the model estimated using transfer learning performs better than both the model estimated on the global dataset and the model estimated on the regional dataset.

\begin{table}
  \caption{MSLE computed on each region's evaluation dataset}
  \label{tab:transfer_evaluation}
  \begin{tabular}{c|ccc}
    \toprule
    Region & Global & Regional & Transfer learning \\
    \midrule
    Southeast Asia & 1.000 & 1.070 & 0.901 \\
    India & 0.375 & 0.508 & 0.351 \\
    Brazil & 0.332 & 0.298 & 0.318 \\
    Australia & 0.311 & 0.434 & 0.305 \\
    Russia & 0.396 & 0.466 & 0.387 \\
    \bottomrule
\end{tabular}
\end{table}

\subsection{Features analysis}

In this section, we analyze how each feature impacts the model prediction by building a partial dependence plot for each feature. The partial dependence plot shows, for each value of a given feature, the average prediction if the feature is set at this value for all the data points. We have used this method because its results are particularly simple to understand. It is worth to keep in mind its limitations, in particular when the features are correlated, which is obviously the case here.

\begin{figure}[h]
  \centering
  \includegraphics[width=\linewidth]{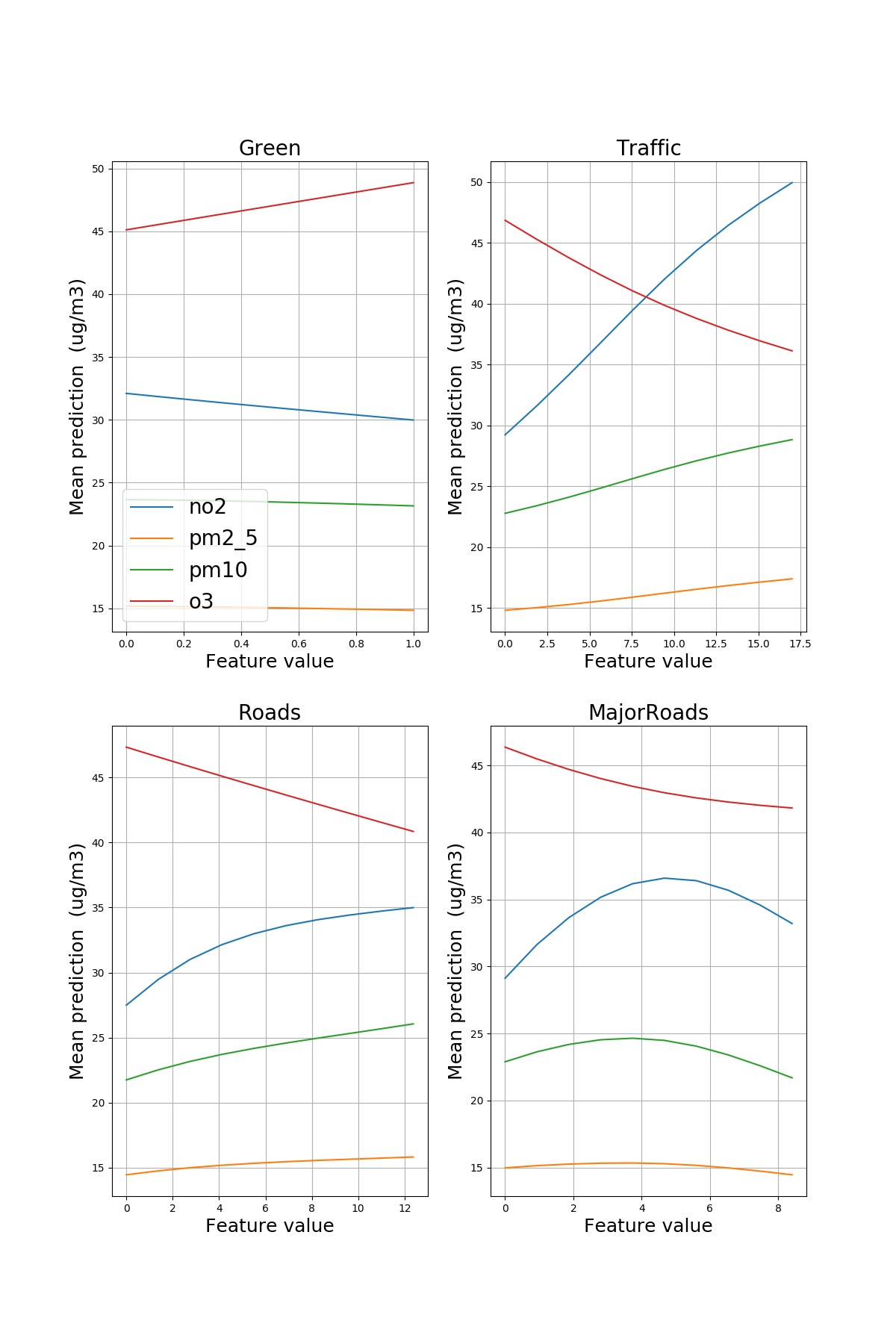}
  \caption{Partial dependence plot for some features in Europe}
  \label{fig:model_sensitivity}
  \Description{Partial dependence plot for some features in Europe}
\end{figure}

Figure~\ref{fig:model_sensitivity} shows the partial dependence plots built for some features using Europe prediction model. Traffic, Roads and MajorRoads features have a high impact on the predictions: higher values are linked to an increase in concentrations for NO$_{2}$, PM$_{2.5}$ and PM$_{10}$ and to a decrease in O$_{3}$ concentration, which is the expected behaviour as the chemical interaction between NO$_{2}$ and O$_{3}$ make them unable to co-exist. The green feature has a more limited impact and a higher value is linked to an increase of O$_{3}$ and to a decrease in NO$_{2}$: this is, again, an expected behaviour as NO$_{2}$ sources are highly related to car exhausts and residential heating.

\section{Applications}

We describe here three applications made possible by this prediction engine: city-level street-level air quality mapping, air quality adjusted routing and large-scale air quality mapping.

\subsection{Street-level air quality mapping}

\begin{figure}[h]
  \centering
  \includegraphics[width=\linewidth]{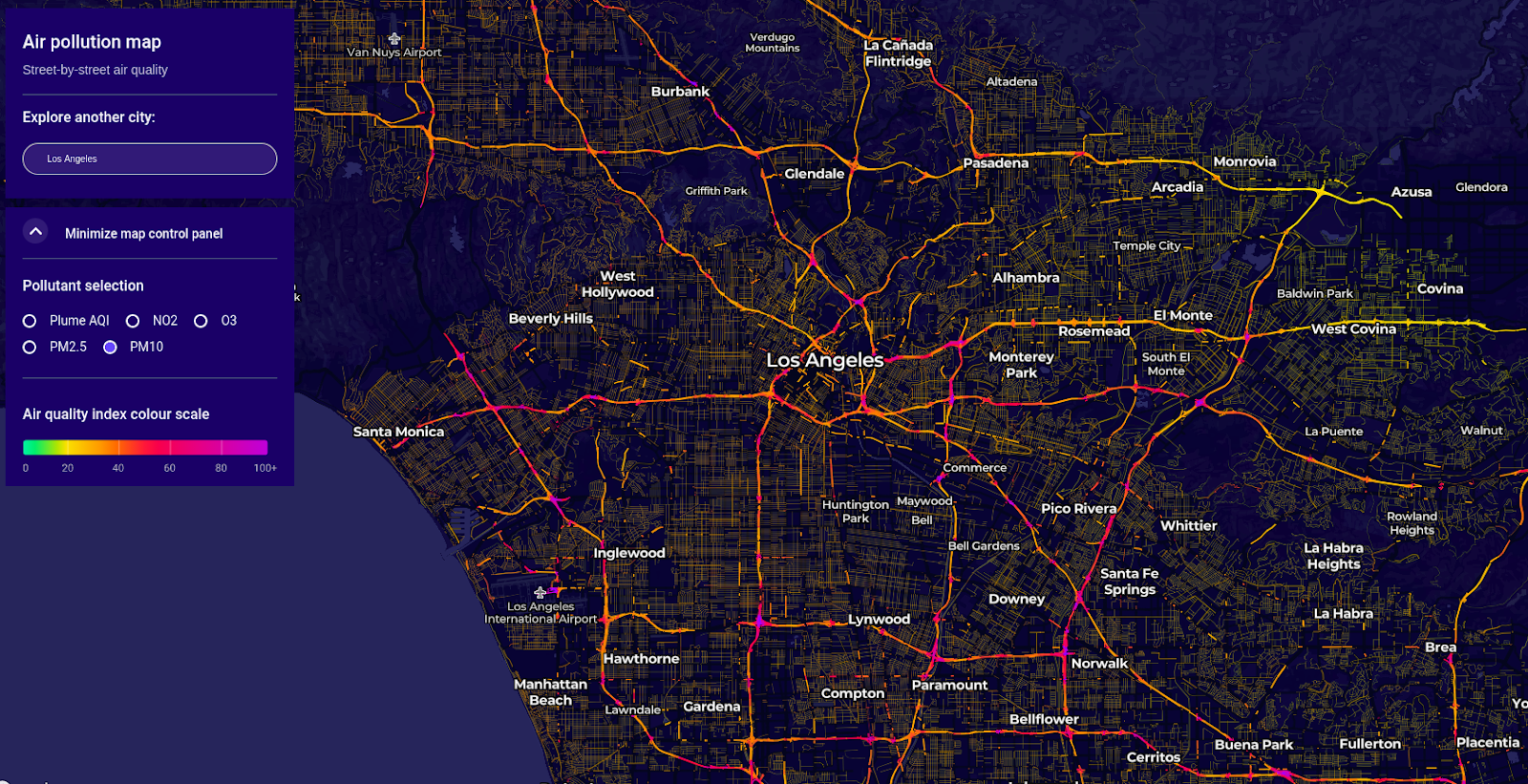}
  \caption{Street-level air quality map in Los Angeles}
  \label{fig:street_level}
  \Description{Street-level air quality map in Los Angeles}
\end{figure}

The prediction engine is used to build real-time street-level air quality maps in the world's largest cities such as the one shown on Figure~\ref{fig:street_level}. This map has been built by applying the prediction engine at every road segment in the city.

\subsection{Air quality adjusted routing}

In a city where street-level air quality mapping has been performed, we can build a routing engine for pedestrians which proposes several paths: the shortest one (what is returned by classical routing engines), but also alternative paths which may be longer but with a lower air pollution exposure.

\begin{figure}[h]
  \centering
  \includegraphics[width=\linewidth]{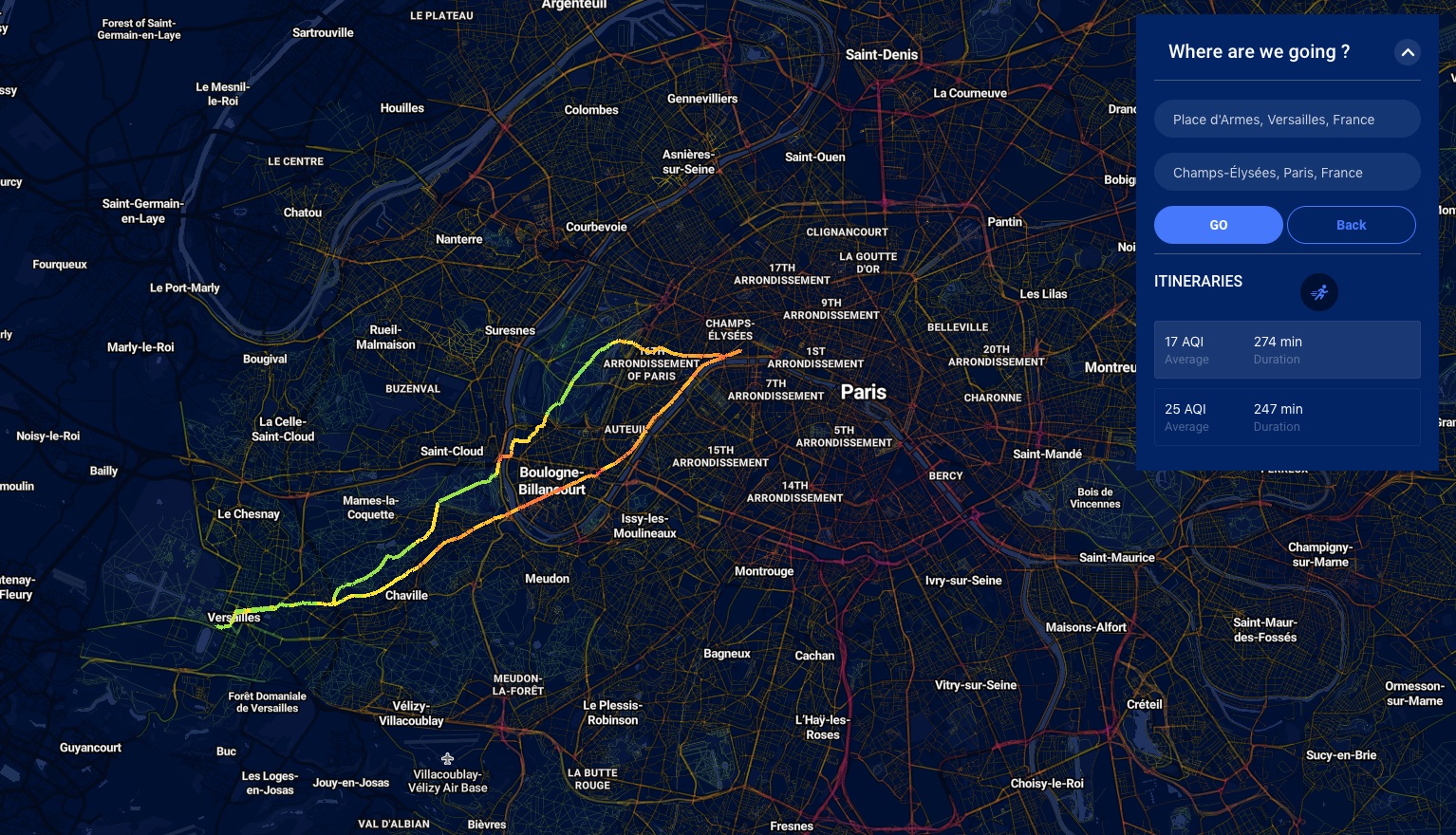}
  \caption{Air quality adjusted routing in Paris}
  \label{fig:routing}
  \Description{Air quality adjusted routing in Paris}
\end{figure}

Figure~\ref{fig:routing} shows an illustration of this routing engine in Paris. The shortest path is obtained by applying a classical shortest path algorithm on the road network where each road segment is weighted by its length. The alternative path is obtained by weighting each road segment by the product of its length and its predicted Plume AQI. In this example, the alternative path is $11\%$ longer but $32\%$ less polluted.

\subsection{Large-scale air quality mapping}

\begin{figure}[h]
  \centering
  \includegraphics[width=\linewidth]{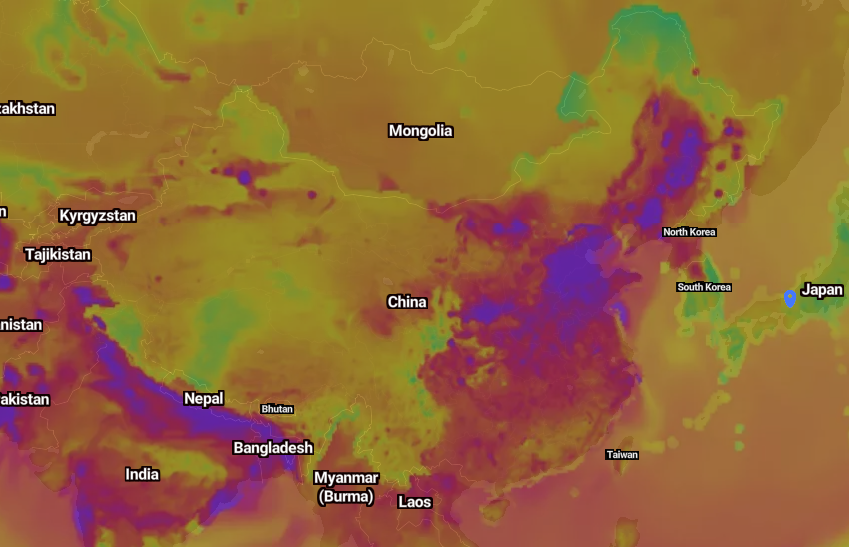}
  \caption{Large-scale air quality map over China}
  \label{fig:heatmap}
  \Description{}
\end{figure}

The prediction engine can also be used to build large-scale air quality heatmaps like the one shown on Figure~\ref{fig:heatmap}. It is worth noting that global air pollution visualizations must be presented using a global air quality index: we choose Plume AQI (presented in \cite{plume_aqi}) which is as far as we know the only one available.

\section{Conclusion and future work}

The engine presented in this paper is at our knowledge the first modeling framework covering such a large part of the world with a resolution of a few dozens of meters. Indeed, most of existing works focus on a paticular city or country with a resolution higher than $1$ kilometer. Reaching such a fine resolution is needed to produce accurate air quality mappings given the high spatial variability of air quality, and enables to build innovative applications that individuals can use to protect themselves from air pollution, like air quality adjusted routing.

A scope for improvement is on how the features feeding the engine are built. In the work described here, they are computed using an isotropic exponential kernel. Defining other types of kernels, and in particular non-isotropic ones, would lead to a better modeling of spatial correlations. The kernels could also be learnt in the way they are learnt in a convolutional neural network, but this would make more difficult to reach a very fine resolution on large regions. Defining and precomputing fixed kernel functions is what enabled us to reach such a high resolution compared to other state-of-the-art papers using convolutional neural networks.

The integration of mobile sensors measurements in an air quality prediction engine is particularly challenging, and we belive that it brings a very significant improvement in the predictions accuracy. These results will be presented in a future paper focused on this topic.

Finally, this paper studies the air quality spatial variability only and can be used to perform real-time air quality mapping. A logical improvement of the engine would be to integrate temporal variability as well to be able to provide temporal forecasts. This makes the prediction problem much more complex and more hardly compatible with the fine resolution reached here.